\documentclass[10pt,twocolumn,letterpaper]{article}

\usepackage{iccv}
\usepackage{times}
\usepackage{epsfig}
\usepackage{graphicx}
\usepackage{amsmath}
\usepackage{amssymb}
\usepackage{pifont}

\usepackage{algorithm}
\usepackage{algorithmic}
\usepackage{setspace}
\usepackage{amsmath,amssymb, bm}
\usepackage{multirow}
\usepackage{tabularx}
\usepackage{diagbox}
\usepackage{booktabs}
\usepackage[mathscr]{euscript}
\usepackage{color}

\usepackage{color}
\usepackage{mathastext}

\usepackage[breaklinks=true,bookmarks=false,colorlinks=true]{hyperref}

\def\iccvPaperID{1762} 

\iccvfinalcopy 

\newcommand{\cmark}{\ding{51}}%
\newcommand{\xmark}{\ding{55}}%

\newcommand*{\affaddr}[1]{#1} 
\newcommand*{\affmark}[1][*]{\textsuperscript{#1}}
\newcommand*{\email}[1]{\texttt{#1}}
\definecolor{orange}{rgb}{1,0.5,0}

\usepackage[normalem]{ulem}

\makeatletter
\newcommand{\printfnsymbol}[1]{%
  \textsuperscript{\@fnsymbol{#1}}%
}
\makeatother

\begin{document}

\title{Proposal, Tracking and Segmentation (PTS): \\ A Cascaded Network for Video Object Segmentation}

\author{
	Qiang Zhou\affmark[1]\printfnsymbol{1}, Zilong Huang\affmark[1]\printfnsymbol{1}, Lichao Huang\affmark[2], Han Shen\affmark[2],  Yongchao Gong\affmark[2], \\
	\and Chang Huang\affmark[2], Wenyu Liu\affmark[1], Xinggang Wang\affmark[1]\\
	\affaddr{\affmark[1]School of EIC, Huazhong University of Science and Technology}\\
	\affaddr{\affmark[2]Horizon Robotics}\\
	\email{\tt\small\{theodoruszq,hzl,xgwang,liuwy\}@hust.edu.cn}\\
	\email{\tt\small{\{lichao.huang,yongchao.gong,han.shen,chang.huang\}@horizon.ai}} \\
}
\maketitle

\begin{abstract}
Video object segmentation (VOS) aims at pixel-level object tracking given only the annotations in the first frame. Due to the large visual variations of objects in video and the lack of training samples, it remains a difficult task despite the upsurging development of deep learning. Toward solving the VOS problem, we bring in several new insights by the proposed unified framework consisting of object proposal, tracking and segmentation components. The object proposal network transfers objectness information as generic knowledge into VOS; the tracking network identifies the target object from the proposals; and the segmentation network is performed based on the tracking results with a novel dynamic-reference based model adaptation scheme. Extensive experiments have been conducted on the DAVIS'17 dataset and the YouTube-VOS dataset, our method achieves the state-of-the-art performance on several video object segmentation benchmarks. We make the code publicly available at \url{https://github.com/sydney0zq/PTSNet}.
\end{abstract}

\section{Introduction}

\let\thefootnote\relax\footnotetext{\textsuperscript{$*$}Equal contributions. The work was mainly done during an internship at Horizon Robotics.}

Video object segmentation (VOS) aims at segmenting specific objects throughout a video sequence, given only the annotations in the first frame. This task is also known as semi-supervised video object segmentation. It has attracted increasing attention due to the availability of large-scale datasets~\cite{davis17_arxiv, yvos_eccv18} and its wide application potential in video editing, autonomous driving \etc. There are several traditional research directions, \eg reduce computational effort~\cite{temporal_superpixels_vos_cvpr13, seamseg_vos_cvpr14, fully_connected_vos_iccv15}, casting the problem into a bilateral space~\cite{bilateral_space_vos_cvpr16}, or considering optical flow~\cite{graph_based_vos_cvpr10, seamseg_vos_cvpr14}. Deep learning based methods mostly originate from OSVOS~\cite{osvos_cvpr17} and MaskTrack~\cite{masktrack_cvpr17}.





Recently some noteworthy methods~\cite{pml_cvpr18, vos_by_lse_eccv18} attempt to solve the VOS problem by pixel-level metric learning. PML~\cite{pml_cvpr18} learns an embedding space via deep metric learning for all labelled pixels of training images; in testing time each pixel of the current frame is assigned to a label by nearest-neighbor matching to the pixels in the first frame. And Hai \etal~\cite{vos_by_lse_eccv18} propose location-sensitive embedding further for distinguish similar instances. In the field of detection, Mask R-CNN~\cite{mask_rcnn_iccv17}, one of the state-of-the-art instance segmentation methods in still images, greatly reduces the searching space of detection and segmentation after introducing \textit{objectness}, which also had been applied for object detection in image and video in many previous works, \eg, \cite{faster_rcnn_iccv15, video_det_tubelet_proposal_cvpr17}. The term objectness means a high-level semantic concept showing whether one region contains objects. As for semi-supervised VOS which extremely lacks of training samples, objectness also has the ability to work in a similar manner in Mask R-CNN and transfer generic semantic recognition. However, objectness has not been exploited in the field of VOS.


Moreover, appearance-based methods like OSVOS~\cite{osvos_cvpr17}, OnAVOS~\cite{onavos_bmvc17} and OSVOS-S~\cite{vos_wo_tempinfo_pami18} rarely depend on temporal consistency, thus less likely to drift when occlusion and abrupt motion occurs. Nevertheless, temporal information is critical for object recognition in video. Propagation-based methods like RGMP~\cite{rgmp_cvpr18} depend on the previous predicted mask and the first annotated frame as temporal and appearance cues respectively. And the two types of VOS methods mix the tasks of localization and segmentation into a single network. 
How about accomplishing the two tasks in two steps?  Our motivation is introducing an object identification mechanism for VOS based on the objectness metioned above. On one hand, it would be relatively straightforward to identify the object of interest in that the searching space of localization is greatly reduced. On the other hand, localization and segmentation desires translation-invariant and translation-equivalent representations respectively~\cite{mask_rcnn_oral_pdf_iccv17}. The object identification idea will be beneficial to avoid the potential mutual exclusion of the two tasks.

Additionally, model adaptation in VOS is not fully exploited compared to VOT (Visual Object Tracking)~\cite{vot_pami16}. PML~\cite{pml_cvpr18} only adds historical embedding features with high confidence to the reference pool. OnAVOS~\cite{onavos_bmvc17} utilizes the predicted masks to fine-tune the segmentation network. Both RGMP and OSMN~\cite{rgmp_cvpr18, Yang2018osmn} extract temporal information only from the previous predicted mask. S2S~\cite{yvos_eccv18} applies ConvLSTM to preserve the spatial information of previous frames in the hidden states of the model, but the effectiveness of model update is hard to be quantified. Yet in VOT, ECO~\cite{eco_cvpr17} groups historical similar samples to update model and MDNet~\cite{mdnet_cvpr16} applies short-long term model update strategy by memorizing historical samples. Unlike OnAVOS, we use dynamic reference instead of online training to do model update.

To bring in the objectness information, the object identification mechanism and the dynamic reference based model adaption scheme, we propose PTSNet, a cascaded network for VOS, which consists of an Object Proposal Network (OPN), an Object Tracking Network (OTN) and a Dynamic Reference Segmentation Network (DRSN). OPN generates proposals near the object of interest. Then OTN identifies the tracked object out and gets aware of its scale. Finally DRSN updates the appearance information over time and utilizes multiple dynamic references to guide the segmentation. Unlike the previous state-of-the-art method DyeNet~\cite{dyenet_eccv18}, PTSNet is a causal system which performs inference online.

More specifically, given a frame in a sequence, OPN firstly generates class-agnostic object proposals for the current frame and filters out the redundant proposals to obtain candidate proposals. This strategy greatly reduces the searching space for further object identification while maintaining true positive regions of interest. Afterwards, OTN identifies the object of interest by giving confidence scores and updates online for adapting to large and fast changes in object appearance. The highest scoring box is selected to extract the region to be segmented. OTN performs box association in objectness level and identifies reliable proposals in a smaller searching space. DRSN segments the current frame with dynamic frames as reference. It not only uses the first static annotated frame for providing reliable information as reference, which is not up-to-the-minute, but also utilizes historical segmentation results as pseudo ground-truths as dynamic reference to guide the current segmentation.

In summary, our main contributions are highligted as follows:
\begin{itemize}
    \item We propose PTSNet which cascades object proposal, tracking and segmentation sub-networks, which is a novel and effective framework for VOS. 
    \item PTSNet brings in some new insights for VOS: objectness as generic knowledge to overcome the problem of lacking training examples, and a robust object localization scheme based on visual tracking for avoiding the conflict of between translation-invariant and translation-equivalent desired by the localization and segmentation tasks respectively. Besides, a new segmentation model adaptation mechanism, \ie, DRSN, is also introduced.
    \item PTSNet achieves top performance on various competitive VOS datasets regardless with or without online fine-tuning, \eg DAVIS'17 and YouTube-VOS.
\end{itemize}

\begin{figure*}[htp]
    	\centering
    	\includegraphics[width=1.0\linewidth]{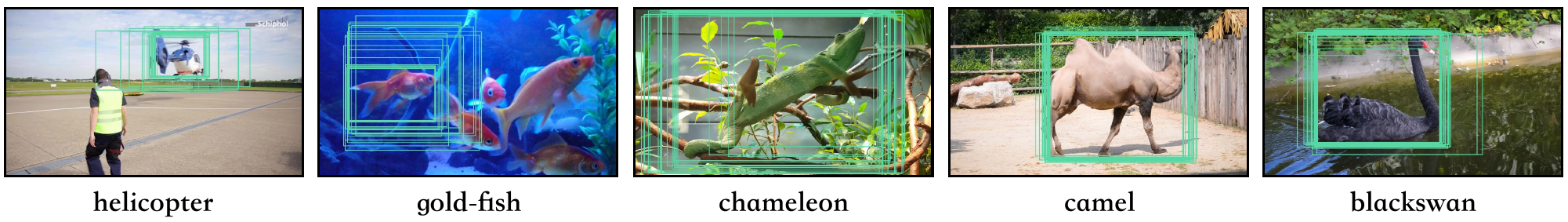}
    	\caption{Example proposals of OPN on unseen categories. We randomly pick a portion of all proposals near the objects of interest. The results show excellent generalization ablity of OPN.}
    	\label{fig:opn_class_agnostic}
\end{figure*}

\section{Related work} \label{Related work}

    \paragraph{Video Object Segmentation.}
    Video object segmentation (VOS) is defined as tracking specified objects in pixel-level given the first annotated frame throughout the video sequence. OSVOS~\cite{osvos_cvpr17} adopts a video-specific segmentation network which learns appearance feature of the target from the first annotated frame to segment the following frames. OnAVOS~\cite{onavos_bmvc17, davis17_arxiv17}, OSVOS-S~\cite{vos_wo_tempinfo_pami18} and CINM~\cite{cnn_in_mrf_cvpr18} extend OSVOS by an mechanism of online adaption,  by learning an semantic instance network and by building a graph model on the results of OSVOS, respectively. MaskTrack~\cite{masktrack_cvpr17} uses optical flow to propagate the segmentation mask from previous frame to the current. LucidTracker~\cite{lucid_datadreaming_arxiv18} extends MaskTrack by a mechanism of data augmentation. RGMP~\cite{rgmp_cvpr18} simultaneously uses both the previously predicted mask to be propagated to the current frame and the first annotated frame as static reference to guide segmentation network to segment the target. MoNet~\cite{monet_cvpr18} introduces two motion exploitation components which are feature alignment and a distance transform layer to achieve better performance. Hu \etal~\cite{motion_guided_vos_cvpr18} proposed an active contour model to provide a coarse segmentation as a guidance cascaded by a refinement network which outputs the final prediction. OSMN~\cite{Yang2018osmn} proposes a modulator network to extract visual information from the first annotated frame and position information from previously predicted mask. S2S~\cite{yvos_eccv18} utilizes a ConvLSTM to learn long-term spatial-temporal information for segmentation. DyeNet~\cite{dyenet_eccv18} combines a bi-directional mask propagation and a ReID module for retrieving missing objects into a single network. \cite{pml_cvpr18, vos_by_lse_eccv18, videomatch_eccv18} explore pixel-level embedding matching for VOS.
    
    Despite achieving impressive results, the above methods disregards the objectness information and most of them does not take translation representations into account. In PTSNet, we propose a principled pipeline to bring in objectness information as generic knowledge, object identification via visual tracking for localization, and an new dynamic model adaptation for segmentation.
    
    \paragraph{Video Object Tracking.}
    In recent works that use deep learning for VOS, the power of VOT has not been fully utilized. FAVOS~\cite{favos_cvpr18} reformulates VOS as a task to track the parts of the target, while we perform object-level tracking. 
    In the field of VOT, two mainstreams of deep-learning based~\cite{siamfc_eccv16, mdnet_cvpr16, siamrpn_cvpr18, goturn_eccv16} and correlation-filter based~\cite{kcf_pami15, eco_cvpr17, c_cot_eccv16} evolved. We applies the high-performance MDNet~\cite{mdnet_cvpr16} which learns shared features using an offline training set and online learning domain-specific classifiers individually for different testing videos. To further speed up the proposed PTSNet, the adaptive RoIAlign in Real-Time MDNet~\cite{rt_mdnet_eccv18} can be introduced. In a word, MDNet acts an object identification module based on the provided objectness information, and performs box association in objectness level for the subsequent segmentation module DRSN. It worth to note that the framework of PTSNet is compatible with most of visual tracking methods and can always benefit from the development of visual tracking. This is also an advantage of PTSNet.

\section{Method} \label{PTSNet}

    \begin{figure*}
    	\centering
    	\includegraphics[width=1.0\linewidth]{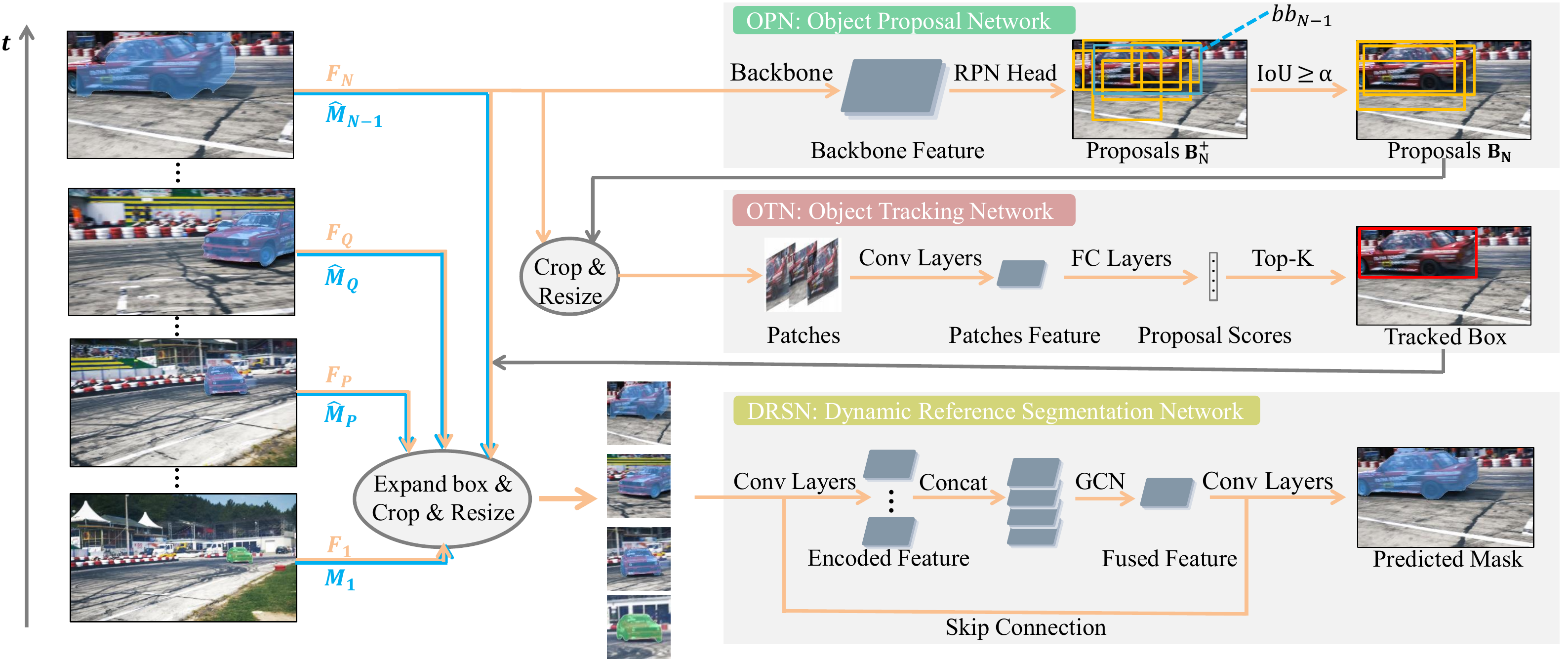}
    	\caption{Overview of the proposed PTSNet, which consists of \textit{Object Proposal Network (OPN)}, \textit{Object Tracking Network (OTN)} and \textit{Dynamic Reference Segmentation Network (DRSN)}. Given a sequence with the annotated first frame ($\bm{F_1}$ and $\bm{M_1}$), PTSNet has already predicted the masks ($\bm{\hat{M}_{2}}$ to $\bm{\hat{M}_{N-1}}$) of $\bm{F_2}$ to $\bm{F_{N-1}}$. Firstly, OPN takes $\bm{F_N}$ as input and generates proposals $\bm{B_{N}^{+}}$, then preserve these proposals near the object of interest by applying IoU threshold with $bb_{N-1}$. These kept proposals are denoted as $\bm{B_{N}}$. Secondly, OTN identifies the best top-K proposals from $\bm{B_{N}}$ to localize the object of interest. Finally, \textbf{\#N} cropped current image with previous mask pair, \textbf{\#Q} cropped image-mask pair, \textbf{\#P} cropped image-mask pair and \textbf{\#1} annotated image-mask pair are fed into the DRSN individually, then we concatenate them and fed it into the subsequent modules to segment the object in the current frame.}
    	\label{fig:overall_detail}
    \end{figure*}

\subsection{Overall framework}

The architecture of the proposed PTSNet is given in Fig.~\ref{fig:overall_detail}.
It consists of
Object Proposal Network (OPN),
Object Tracking Network (OTN)
and Dynamic Reference Segmentation Network (DRSN).
OPN is pre-trained on the COCO dataset~\cite{lin2014microsoft} and provides high-quality proposals near the object of interest. OTN is designed to identify the best proposal and to be updated online for adapting to large and fast changes in object appearance, which is inspired by MDNet~\cite{mdnet_cvpr16}. Then, the best proposal is expanded to crop and resize the region of the object of interest for normalizing scale of the object. Finally, DRSN makes use of both cropped region with previously predicted mask and multiple reference frames to segment the target object in the current frame. 

\subsection{Object Proposals Network} \label{OPN}
Object Proposal Network (OPN) is introduced to generate high-quality proposals near the object of interest in each frame to bring in objectness.

The proposed OPN works in a progressive way: Given the object location, \ie, bounding box, denoted as $bb_{N-1}$, in frame $N-1$, OPN aims at proposing a small set of bounding boxes, denoted as $BB_N = \{\hat{bb}_N^i, i=1, \cdots, k\}$, in frame $N$, as the potential locations of the target object. We follow the basic assumption that the boxes in $BB_N$ should be close to $bb_{N-1}$.

OPN is based on the state-of-the-art region proposal network (RPN) in Mask R-CNN~\cite{mask_rcnn_iccv17} pretrained on the large-scale COCO dataset. RPN is class-agnostic and provides objectness information as generic knowledge for PTSNet. Since VOS aims at segmenting object from any semantic category but COCO only provides the annotation for the 80 categories, it is important to check whether the pretrained RPN has a good generalization ability on unseen categories. Thus, we test RPN on the COCO validation set and VOS datasets (the DAVIS'17 dataset and the YouTube-VOS dataset)~\footnote{For the details of the dataset, please refer to Sec.~\ref{Datasets}} and report the recall rates of RPN. The COCO validation images are not used for training RPN but they are in the same domain with the training images of RPN; the images in the VOS datasets are not in the same domain with the training images of RPN and contain objects from unseen categories. Table~\ref{table:recall} shows that the recall rates of RPN on VOS datasets are higher with the recall rates on the COCO validation set. The results confirm the amazing generalization ability of RPN. Several qualitative results for some unseen categories are shown in Fig.~\ref{fig:opn_class_agnostic}.

    \begin{table}
    \caption{Comparison of recall rates between several datasets in different domains. ``DET" denotes detection while ``VOS" denotes video object segmentation. The recall ratio are obtained by RPN tailored from Mask R-CNN pretrained on the COCO training dataset. It can be observed that RPN achieves higher recall ratio in VOS domain datasets than in DET dataset.}
    \vspace{3mm}
      \centering
      \setlength{\tabcolsep}{2mm}
      {\begin{tabular}{c | c c }
        \toprule
        Dataset                                 & Domain        & Recall (IoU $\geq$ 0.5)    \\
        \midrule
        COCO~\cite{lin2014microsoft}            &  DET          & 83$\%$                    \\ 
        DAVIS'17~\cite{davis_eval_cvpr16}       &  VOS          & 85$\%$                    \\ 
        YouTube-VOS~\cite{yvos_eccv18}          &  VOS          & 87$\%$                    \\ 
        \bottomrule
      \end{tabular}}
      \label{table:recall}
    \end{table}

Based on the high recall rates of RPN on the VOS datasets, we design a hybrid strategy for OPN. Given frame $N$, we denote the output boxes of RPN is denoted as $BB^+_N$. We calculate the box IoU for every box in $BB^+_N$ between $bb_{N-1}$, and then keep the boxes whose IoU are larger than $\alpha$, \ie $0.3$. The kept boxes are $BB_N$. However, since RPN has a probability (13\% $\sim$ 15\%) of failing to localize the object of interest. Thus, if number of elements in $BB_t$ is less than $5$, we fill up $BB_t$ with extra boxes generated by the box sampling method described in~\cite{mdnet_cvpr16}, which is based on the assumption that the location variations of object will not be too large in adjacent frames.

\subsection{Object Tracking Network} \label{OTN}
OPN provides a few proposals near the object of interest, but they are class-agnostic. In this subsection, we aims at building an instance-specific appearance model based on labeled object in the first frame to identify the object of interest throughout the video. This is a reduced visual tracking problem. We call it a ``reduced" tracking problem, since the candidate object locations are already given by OPN. Object Tracking Network (OTN) is introduced to solve this problem.

Our OTN is based on the high-performance deep learning based tracker MDNet~\cite{mdnet_cvpr16}. The differences between OTN and MDNet are (1) OTN is based on OPN while MDNet has its own tracking candidate generation method and (2) they are trained using different datasets. For self-contained of our paper, we introduce the details of OTN/MDNet in details as follows.

Given $BB_N$, we crop its patches and resize them into the same size, \eg, $H \times W$. Afterwards, these patches are passed into a network pretrained on the DAVIS'17 dataset~\footnote{For the details of the datasets, please refer to Sec.~\ref{Datasets}} to predict the confidence scores belonging to the target object for each patch. Then, we draw the top-5 scoring proposals. After that, we compute the mean value of the top-5 scores to check whether the tracking is a success, \ie, the mean value is bigger than 0, a success, otherwise a failure. If succeed, we collect deep features of the patches with spatial confidences into a memory stack. And we average the coordinates of the top-5 proposals as the current prediction. If failed, we do not collect any samples and keep the previous predicted box as the current prediction. Finally, we employ a short-long term model update strategy: If it is a success, we do short term model update by fine-tuning some specified fully connected layers using the collected features in a small window. If it is a failure, we check if it is the time to do regularly long-term model update which draws the whole collected samples to fine-tune the specified layers.

In our experiment, we notice that the collaboration of OPN and OTN results in a more robust object tracker, which significantly alleviates the the burden of subsequent segmentation module. Consequently, the segmentation module only needs to care about the task of pixel-wise classification.

\subsection{Dynamic Reference Segmentation Network} \label{DRSN}
Since coarse localization of target is obtained by the proposed coupled OPN and OTN, the Dynamic Reference Segmentation network (DRSN) is proposed to perform object segmentation with online adaption. Previous reference-guide models~\cite{rgmp_cvpr18, Yang2018osmn} statically refer to the feature from the the first annotated frame. Thus, they are not able to adapt the appearance variations through a video sequence.



    \paragraph{The dynamic inputs of DRSN.}
    The network architecture of DRSN inherits from~\cite{rgmp_cvpr18}. And the main difference is our dynamic reference inputs described as follows: It inputs several (here we use 4 for example) image-mask pairs which consist of the current frame $\bm{F_N}$ and previously predicted mask $\bm{\hat{M}_{N-1}}$, the first frame $\bm{F_1}$ and its annotated mask $\bm{M_1}$, the P-th, Q-th frame $\bm{F_P}$, $\bm{F_Q}$ and its predicted mask $\bm{\hat{M}_P}$, $\bm{\hat{M}_Q}$. In detail, each image-mask pair is cropped from the origin image-mask and resized to the same shape of $4 \times H \times W$. Four channels are composed of three channels of RGB image, and one channel of mask. The choice of P and Q is worth discussing, there are various sampling styles like drawing predictions of highest confidence, equally interval spacing or key frames. In practice, we find equally spaced sampling in a recent time window works well, \ie we could select frame $\mathbf{\max (1, N-4)}$ and $\mathbf{\max (1, N-2)}$ when window size and interval is set to 4 and 2, respectively. Note that in Fig~\ref{fig:overall_detail}, we expand the boxes enclosing their corresponding masks by a constant in height and width, \eg, 1.5. Then we crop the patches from frame $\bm{F_1}, \bm{F_P}, \bm{F_Q}$ by the expanded boxes. Different from them, the image-mask patch of the current frame $\bm{F_N}$ is cropped by the expanded box provided by OTN.

    \paragraph{The network architecture of DRSN.}
    Firstly the four image-mask patch pairs are fed into an convolutional network, \eg, ResNet-50~\cite{he2016deep}, to extract deep encoded features. Then the four features are concatenated in channel axis and fed into a global convolution network (GCN)~\cite{peng2017large} and three refinement modules~\cite{refinement_module_eccv16}. It is expected that the GCN could properly aggregate the information from the labeled first frame and the recent appearance variations, and treat them as the supplementary information to help the prediction of current frame. The refinement modules merge features in three scales ($1/4, 1/8, 1/16$ shape of the input image-mask patch) and produce the output ($1/4$) with two channels indicating the probability of foreground and background. Finally bilinear interpolation is applied for the final mask prediction. 
\\

Unlike OnAVOS~\cite{onavos_bmvc17}, DRSN performs online adaptation without fine-tuning the network with its predictions. The first annotated frame donated as static reference can always provide reliable but not up-to-the-minute appearance feature, which is the opposite of the appearance features generated from frames nearby donated as dynamic reference. Although the predicted masks may be inaccurate, it still offers effective appearance cues of the object of interest. In practice, to balance efficiency and performance, we choose the first annotated frame and another two historical frames close to the current frame and as reference. The proposed DRSN takes the advantage of both the static and dynamic references to obtain better performance.

\section{Experiments} \label{Experiments}
In this section, we will firstly introduce the datasets and our implementation details, then compare PTSNet to state-of-the-art methods. At last, ablation and add-on studies for each component of PTSNet will be revealed to validate their effectiveness.

\subsection{Datasets and evaluation metric} \label{Datasets}

    \paragraph{Datasets.}
    We evaluate the proposed PTSNet on the DAVIS'17~\cite{davis_eval_cvpr16} dataset and the recently released YouTube-VOS~\cite{yvos_eccv18} dataset. The two datasets contain assorted challenges, such as appearance and pose variations, motion blur and object occlusion.
    
    DAVIS dataset has two sets, DAVIS'16 and DAVIS'17. The differences of the two sets are the number of sequences and whether multiple instances are annotated in each sequence. Here we choose the latter one yet the harder one for our experiments. DAVIS'16 only separates foreground and background for each frame so that there is no objectness concept when appearing multiple instances. However, the major functionality of PTSNet aims at video object segmentation. This is the reason why we do not conduct experiments on DAVIS'16. DAVIS'17 consists of 60 sequences for training and 30 sequences for evaluation, totally 8294 frames. Each frame is provided with pixel-level annotations, where one single instance or multiple different instances are separated from the background.
    
    YouTube-VOS dataset is the largest dataset for VOS so far. It contains 4453 sequences and is split into training (3471), online validation (474) and online testing (508) sets. The training set is annotated for every 5 frames, comprising one single instance or multiple different instances just like DAVIS'17. It is noteworthy that the online validation set accepts predictions of every 5 frames while the online testing set requires predictions of each frame.
    
    \paragraph{Evaluation metric.}
    For DAVIS'17 and YouTube-VOS dataset, we follow~\cite{davis_eval_cvpr16} that adopts region similarity ($\mathcal{J}$), contour accuracy ($\mathcal{F}$) and their average ($\mathcal{G}$) measures for evaluation. Region similarity ($\mathcal{J}$) is calculated as the average IoU between the proposed masks and the groundtruth masks respectively, while the contour accuracy ($\mathcal{F}$) interprets the proposed masks as a set of closed contours and computes the contour-based F-measure which is a function of precision and recall. Generally speaking, region similarity measures the ratio of correctly labeled pixels predicted by algorithms and contour accuracy measures the precision of the segmentation boundaries.

\subsection{Implementation details}
As mentioned in Sec.~\ref{PTSNet}, the proposed PTSNet is composed of OPN (Sec.~\ref{OPN}), OTN (Sec.~\ref{OTN}) and DRSN (Sec.~\ref{DRSN}), whose details are described as follows respectively.

\paragraph{OPN.} It is based on RPN tailored from Mask-RCNN~\cite{mask_rcnn_iccv17}. We use the ResNeXt-152~\cite{Xie2016} backbone edition pretrained on the COCO dataset. The weights of RPN are fixed and NMS (Non-maximum suppression) and threshold restriction are removed to obtain about 2000 proposals for each frame. The $\alpha$ of filtering out irrelevant proposals is set to $0.3$. If RPN fails to localize the object of interest, we draw 256 samples and the spatial and scale hyper-parameters of Gaussian box sampling~\cite{mdnet_cvpr16} are set to 0.1 and 1.5. 


\paragraph{OTN.} It inherits MDNet~\cite{mdnet_cvpr16} and is trained on DAVIS, based on the ImageNet pretrained VGG-M~\cite{chatfield14return} backbone. Considering the success of MDNet with few training videos, it is enough to train a high performance tracker. And the recent Real-Time MDNet~\cite{rt_mdnet_eccv18} shows that it does not gain much improvement after switching to a larger dataset. The hyper-parameters of OTN are almost same with MDNet. The main differences are (1) The final bounding box regression is removed as we have high-quality proposals already, because the videos in DAVIS dataset are usually shorter than the ones in VOT datasets and appearance variations over time are usually larger (2) We have cut down the updating window sizes of short term (from 20 to 5) and long term (100 to 20).

\paragraph{DRSN.} The weights of encoder is initialized by ResNet-50 pretrained on ImageNet. For the evaluation of YouTube-VOS, we only train DRSN on YouTube-VOS for 100k iteration and batch size of 64. Adam~\cite{kingma2014adam} optimizer, the initial learning rate of 2e-05 (decay by 0.1 after 60k iterations) on 4 NVIDIA Titan V GPUs. The shape of image-mask pair is $4 \times 256 \times 256$ and the channels of GCN and refinement modules are all 256. For the evaluation of DAVIS'17, we take the model pretrained on YouTube-VOS, then continue training on DAVIS'17 training set for an extra 35k iterations, with a learning rate of 2e-07 (decay to 2e-8 after 12.5k iterations).

In the training phase, we firstly pick up a target from an arbitrary sequence and respectively its 3 sampling frames, as mentioned in section~\ref{DRSN}. A random shift similar to ~\cite{goturn_eccv16} is applied to each enclosed box of the ground-truth mask, so as to simulate the characteristic of boxes predicted by OTN. Afterwards, we enlarge the shifted boxes by 1.5 times along the width and height axes, in order to ensure the completeness of the object in patch. Next, the four stacked tensors, including an image-mask pair of the sampled first frame, two image-mask pairs of the sampled P-th and Q-th frames, along with a pair of image in N-th and blurred mask simulating the previous prediction, are fed into DRSN. The loss function is defined as the standard pixel-wise cross entropy to measure the similarity of the final predicted mask and the groundtruth mask. We also use data augmentation strategies, like mirror flipping and illumination enhancement to increase the robustness of our model.

As for online fine-tuning, we randomly generate image pairs and their masks only from the annotated first frame, keeping the same settings in training stage. The model is fine-tuned on these pairs for 400 iterations with a initial learning rate of 1e-06 with an Adam optimizer.


\subsection{The DAVIS'17 benchmark} \label{exp_davis}

    \begin{table}[b]
    \caption{Quantitative evaluation of our method compared with the state-of-the-art results from previous literature on the DAVIS'17 validation set. For each method, we report whether it employs online fine-tuning (OF) using the first frame as defined in \cite{osvos_cvpr17}, is it causal, and the final performance $\mathcal{J}$ Mean, $\mathcal{F}$ Mean and $\mathcal{G}$. Without OF and under the restriction of causality, our approach surpasses current state-of-the art methods. Further, our approach performs better than DyeNet when applied online fine-tuning but still kept causal.}
    \vspace{1mm}
      \centering
      \setlength{\tabcolsep}{1.2mm}
      {\begin{tabular}{c | c c | c c c}
        \toprule
        Method & OF & Causal & $\mathcal{J}$ Mean & $\mathcal{F}$ Mean & $\mathcal{G}$ \\
        \midrule
        OSVOS~\cite{osvos_cvpr17}                 & \cmark & \cmark &      56.6     &       63.9       &        60.3 \\
        OnAVOS~\cite{onavos_bmvc17}               & \cmark & \cmark &      64.5     &       71.2       &        67.9 \\
        OSVOS-S~\cite{vos_wo_tempinfo_pami18}     & \cmark & \cmark &      64.7     &       71.3       &        68.0 \\
        CINM~\cite{cnn_in_mrf_cvpr18}             & \cmark & \cmark &      67.2     &       74.2       &        70.6 \\
        OSMN~\cite{Yang2018osmn}                  & \xmark & \cmark &      52.5     &       57.1       &        54.8 \\
        RGMP~\cite{rgmp_cvpr18}                   & \xmark & \cmark &      64.8     &       68.6       &        66.7 \\
        VideoMatch~\cite{videomatch_eccv18}       & \xmark & \cmark &      61.4     &       -          &        -    \\
        FAVOS~\cite{favos_cvpr18}                 & \xmark & \cmark &      54.6     &       61.8       &        58.2 \\
        DyeNet~\cite{dyenet_eccv18}               & \xmark & \xmark &      -        &       -          &        74.1 \\
        PReMVOS~\cite{prevmvos_accv18}            & \cmark & \xmark &      77.8     &       73.9       &        81.7 \\
        \midrule
        \textbf{PTSNet(Ours)}                     & \xmark & \cmark &      66.1     &       70.5       &        68.3 \\
        \textbf{PTSNet(Ours)}                     & \cmark & \cmark &      71.6     &       77.7       &        74.7 \\
        \bottomrule
      \end{tabular}}
      \label{table:davis17}
    \end{table}

    \begin{figure*}[!t]
    	\centering
    	\includegraphics[width=1.0\linewidth]{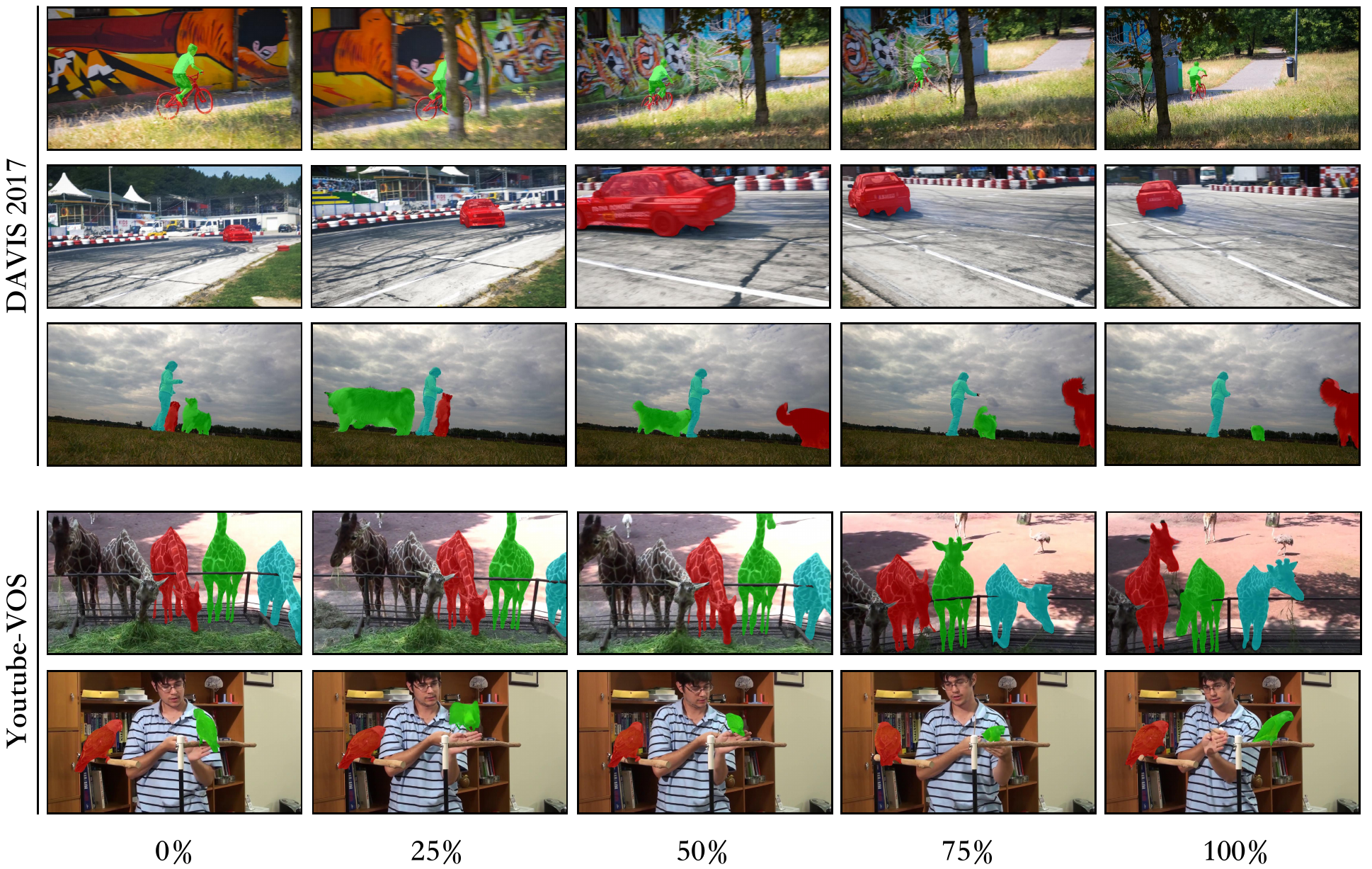}
    	\caption{Visualization results uniformly sampled in videos on the DAVIS'17 and YouTube-VOS datasets. The ``$0\%$" column presents the annotated first frames.}
    	\label{fig:visual_res}
    \end{figure*}

In Table~\ref{table:davis17}, the quantitative results of our approach are compared with other methods in previous literature, and our method achieves top performance. In the circumstance of discarding online fine-tuning (OF), PTSNet outperforms many state-of-the-art methods like OSMN, RGMP and VideoMatch except DyeNet and PReMVOS. However, the inference setting of DyeNet and PReMVOS is iterative inference which makes the method non-casual. It is unfair for the causal methods to compare with the methods with iterative inference. In the circumstance of equipped with online fine-tuning, PTSNet outperforms all listed state-of-the-art methods by a large margin, and it is even better than DyeNet. The performance of PReMVOS is higher than ours with online fine-tuning, however our method is free of burdened modules, \eg ReID and FlowNet.

\subsection{The YouTube-VOS benchmark}
YouTube-VOS is a recently released largest-scale dataset for VOS. The validation set contains 474 sequences with 65 seen classes in training set and 26 classes which are not included. We compare our results with previous published literature~\cite{yvos_eccv18, appearance_model_for_end2end_vos_arxiv1811}. Our results are obtained by submitting to the official evaluation server.

    \begin{table}[b]
    \caption{Comparisons with start-of-the-art methods on the YouTube-VOS validation set. $\mathcal{J}$ denotes the region similarity and the contour accuracy. ``s" and ``u" denote the results averaged over the seen categories and unseen categories, respectively. ``Mean" denotes the results averaged over $\mathcal{J}$ and $\mathcal{F}$. ``OF" denotes online fine-tuning.}
    \vspace{2mm}
      \centering
      \setlength{\tabcolsep}{2mm}
        {\begin{tabular}{c | c | c c | c}
        \toprule
        Method                              &   OF    & $\mathcal{J}$ s & $\mathcal{J}$ u & $\mathcal{G}$ Mean \\
        \midrule
        OSVOS~\cite{osvos_cvpr17}           & \cmark  & 59.8 & 54.2 & 58.8 \\
        OnAVOS \cite{onavos_bmvc17}         & \cmark  & 60.1 & 46.6 & 55.2 \\
        S2S \cite{yvos_eccv18}              & \cmark  & 71.0 & 55.5 & 64.4 \\
        \midrule
        MaskTrack~\cite{masktrack_cvpr17}   & \xmark  & 59.9 & 45.0 & 53.1 \\
        OSMN \cite{Yang2018osmn}            & \xmark  & 60.0 & 40.6 & 51.2 \\
        S2S \cite{yvos_eccv18}              & \xmark  & 66.7 & 48.2 & 57.6 \\
        RGMP \cite{rgmp_cvpr18}             & \xmark  & 59.5 & 45.2 & 53.8 \\
        \midrule
        \textbf{PTSNet(Ours)}               & \xmark  & 69.1 & 53.5 & 63.2 \\
        \textbf{PTSNet(Ours)}               & \cmark  & 73.5 & 64.3 & 71.6 \\
        \bottomrule
      \end{tabular}}
      \label{table:yvos18}
    \end{table}

We also present some state-of-the-art results from literature on the YouTube-VOS official validation set in Table~\ref{table:yvos18}, the proposed PTSNet significantly outperforms the other methods~\cite{osvos_cvpr17, masktrack_cvpr17, Yang2018osmn, onavos_bmvc17, yvos_eccv18} under the circumstances of with or without online fine-tuning. 

Figure~\ref{fig:visual_res} shows some successful video object segmentation results. It indicates PTSNet is robust to occlusion (row 1 and 3), large scale and appearance change (row 2), camera motion (row 4) \etc. However, PTSNet has the risk of drifting in case of long-term large occlusion. This is mainly because after long-term occlusion, OPN have no elaborate means to keep the proposals near the object of interest when the object appears again. Further speaking, the assumption that the object of interest will not move too much within nearby frames can not be utilized in such circumstance. And when the object reappears after long-term occlusion, OTN has no proper mechanism to find out the precise location of the object, which leads to the failure of PTSNet in the end.

\subsection{Ablation study} \label{abstudy}

To demonstrate the effectiveness of the three proposed components, we have designed several groups of ablation studies. All the experiments are conducted on the DAVIS'17 validation set and the performance is measured by $\mathcal{J}$ Mean. Besides, we also evaluate DAVIS as a tracking dataset to prove the effectiveness of objectness information.

    
    \begin{table}[!t]
    \caption{Ablation studies of PTSNet on the DAVIS'17 validation set, measured by $\mathcal{J}$ Mean. We use RGMP as our baseline for segmentation. Dy-Ref. denotes dynamic reference. OTN and OPN denote object tracking network and object proposal network respectively. OF denotes online fine-tuning on the first frame.}
    \vspace{1mm}
      \centering
      \setlength{\tabcolsep}{1mm}
      {\begin{tabular}{c c c c c | c c}
        \toprule
        RGMP        & Dy-Ref.   & OTN       & OPN       & OF        & $\mathcal{J}$ Mean & $\Delta$  \\
        \midrule
        \cmark      &           &           &           &           &   58.8   &  \\
        \cmark      & \cmark    &           &           &           &   63.9   & +5.1 \\
        \cmark      & \cmark    & \cmark    &           &           &   64.3   & +5.5 \\
        \cmark      & \cmark    & \cmark    & \cmark    &           &   66.1   & +7.3 \\
        \cmark      & \cmark    & \cmark    & \cmark    & \cmark    &   71.6   & +13.6 \\
        \bottomrule
      \end{tabular}}
      \label{table:ablationstudy}
    \end{table}
    
    \begin{table}[b]
    \caption{Study the effectiveness of OPN by evaluating DAVIS as a tracking dataset. ``G-sampling" denotes the Gaussian sampling proposed in MDNet.~\cite{mdnet_cvpr16} Both entries use a same OTN exactly and the only difference is the module of proposal generator.}
    \vspace{1mm}
      \centering
      \setlength{\tabcolsep}{1mm}
      {\begin{tabular}{c c c | c c}
        \toprule
        G-Sampling              & OPN       & OTN       & AUC       & $\Delta$  \\
        \midrule
        \cmark                  &           & \cmark    & 52.1      &           \\
                                & \cmark    & \cmark    & 77.8      &   +25.7    \\
        \bottomrule
      \end{tabular}}
      \label{table:davis_vot}
    \end{table}

\paragraph{Single reference \vs Dynamic reference.} As shown in row 1 and 2 in Table~\ref{table:ablationstudy}, we compare two reference settings of the segmentation network, which are static reference and our dynamic reference, named RGMP and DRSN respectively. Note that we disregard the BPTT~\cite{rgmp_cvpr18} trick proposed in the original paper for fair comparison, and the training dataset and hyper-parameter are kept same as well. It shows that DRSN outperforms RGMP by $5.1$, indicting dynamic reference significantly boosts performance. We also analyze the ideal number of reference frames and consuming time (Titan V) used by DRSN, which is shown in Table~\ref{table:drsn_ref_as}. To balance efficiency and speed, we select the setting in the third row with $66.1$ $\mathcal{J}$ Mean and $56.15$ ms per frame.

\paragraph{The effectiveness of objectness clues.} We analyze the influence by whether introducing OPN respectively. Several experiments are conducted to test our assumption.
\\

DRSN w/ OTN (row 3) uses simple Gaussian sampling the same as what MDNet does without objectness information involved. The reliability of tracking is determined by the straight criterion mentioned in the last passage. As shown in Table~\ref{table:ablationstudy}, it achieves a margin gain of $0.4$ (row 3 \vs row 2).

DRSN w/ OPN and OTN (row 4) gets a notable $2.2$ improvement from $\mathcal{J}$ Mean of $63.9$ to $66.1$ (row 3 \vs row 4) by adding OPN module, which illustrates the fact that the object-level guidance indeed helps mask tracking compared to the experiment in last paragraph. By the way, online fine-tuning boosts our methods to $71.6$ $\mathcal{J}$ Mean further.

    \begin{table}[!t]
    \caption{Study of the ideal number of reference frames. ``Gt" means the given first frame; ``Ref (i)" denotes the frame index as the dynamic reference. ``N" denotes the index of the current frame. The speed is tested on Titan V.}
    \vspace{3mm}
      \centering
      \small
      \setlength{\tabcolsep}{2mm}
      {\begin{tabular}{c c c c | c c}
        \toprule
        Gt & Ref (1) & Ref (2) & Ref (3) & $\mathcal{J}$ Mean & Time (ms) \\
        \midrule
        1 &  -  &  -  & -   & 63.0 & \textbf{33.79} \\
        1 &  -  &  -  & N-2 & 64.9 & 43.15 \\
        1 &  -  & N-4 & N-2 & \textbf{66.1} & 56.15 \\
        1 & N-6 & N-4 & N-2 & 66.0 & 70.01 \\
        \bottomrule
      \end{tabular}}
      \label{table:drsn_ref_as}
    \end{table}

\paragraph{Evaluating DAVIS as a tracking dataset.}
    To prove the effectiveness of introduction of OPN by qualitative analysis, we treat the DAVIS'17 validation set as a tracking dataset. Before the evaluation, we firstly train OTN on DAVIS'17 training dataset by extracting tracking boxes from masks. Then we use the same OTN to evaluate: (a) A Gaussian proposal generator proposed in MDNet stacked by OTN (b) OPN stacked by OTN. We use standard AUC~\cite{otb2013} metric as the criterion. In Table~\ref{table:davis_vot}, the impressive performance enhancement shows that the introduction of objectness is a key factor in our framework.

\section{Conclusion and future work} \label{Conclusion}
In this paper, we propose PTSNet, a cascaded framework for semi-supervised VOS. Our PTSNet reaches the current state-of-the-art performance in an intuitive way, bringing in objectness and tracking from object level to pixel level. Owning to the decomposing video object segmentation into three sub-modules (Object Proposal Network, Object Tracking Network, Dynamic Reference Segmentation Network), PTSNet can handle large scale and appearance variations, respectively. With the modular design, PTSNet can easily benefit from other state-of-the-art methods to achieve scalable performance. 

There still remains many future directions in our framework. For example, we can integrate modules in a more elegant way to enable end-to-end training. To make PTSNet more robust on challenging scenes with long time object occlusion or crowded objects, re-identification module could be used for long-term association to alleviate problems such as lost track caused by occlusion or ID switch between similar objects. 


{\small
	\bibliographystyle{ieee}
	\bibliography{PTS}

\begin{thebibliography}{10}\itemsep=-1pt

\bibitem{seamseg_vos_cvpr14}
S.~Avinash~Ramakanth and R.~Venkatesh~Babu.
\newblock Seamseg: Video object segmentation using patch seams.
\newblock In {\em Proceedings of the IEEE Conference on Computer Vision and
  Pattern Recognition}, pages 376--383, 2014.

\bibitem{cnn_in_mrf_cvpr18}
L.~Bao, B.~Wu, and W.~Liu.
\newblock Cnn in mrf: Video object segmentation via inference in a cnn-based
  higher-order spatio-temporal mrf.
\newblock In {\em IEEE Conference on Computer Vision and Pattern Recognition},
  pages 5977--5986, 2018.

\bibitem{siamfc_eccv16}
L.~Bertinetto, J.~Valmadre, J.~F. Henriques, A.~Vedaldi, and P.~H. Torr.
\newblock Fully-convolutional siamese networks for object tracking.
\newblock In {\em European conference on computer vision}, pages 850--865,
  2016.

\bibitem{osvos_cvpr17}
S.~Caelles, K.-K. Maninis, J.~Pont-Tuset, L.~Leal-Taix{\'e}, D.~Cremers, and
  L.~Van~Gool.
\newblock One-shot video object segmentation.
\newblock In {\em IEEE Conference on Computer Vision and Pattern Recognition},
  pages 5320--5329, 2017.

\bibitem{temporal_superpixels_vos_cvpr13}
J.~Chang, D.~Wei, and J.~W. Fisher.
\newblock A video representation using temporal superpixels.
\newblock In {\em Proceedings of the IEEE Conference on Computer Vision and
  Pattern Recognition}, pages 2051--2058, 2013.

\bibitem{chatfield14return}
K.~Chatfield, K.~Simonyan, A.~Vedaldi, and A.~Zisserman.
\newblock Return of the devil in the details: Delving deep into convolutional
  nets.
\newblock In {\em British Machine Vision Conference}, 2014.

\bibitem{pml_cvpr18}
Y.~Chen, J.~Pont-Tuset, A.~Montes, and L.~Van~Gool.
\newblock Blazingly fast video object segmentation with pixel-wise metric
  learning.
\newblock In {\em IEEE Conference on Computer Vision and Pattern Recognition},
  pages 1189--1198, 2018.

\bibitem{favos_cvpr18}
J.~Cheng, Y.-H. Tsai, W.-C. Hung, S.~Wang, and M.-H. Yang.
\newblock Fast and accurate online video object segmentation via tracking
  parts.
\newblock In {\em IEEE Conference on Computer Vision and Pattern Recognition},
  pages 7415--7424, 2018.

\bibitem{vos_by_lse_eccv18}
H.~Ci, C.~Wang, and Y.~Wang.
\newblock Video object segmentation by learning location-sensitive embeddings.
\newblock In {\em Proceedings of the European Conference on Computer Vision
  (ECCV)}, pages 501--516, 2018.

\bibitem{eco_cvpr17}
M.~Danelljan, G.~Bhat, F.~Shahbaz~Khan, and M.~Felsberg.
\newblock Eco: Efficient convolution operators for tracking.
\newblock In {\em CVPR}, 2017.

\bibitem{c_cot_eccv16}
M.~Danelljan, A.~Robinson, F.~S. Khan, and M.~Felsberg.
\newblock Beyond correlation filters: Learning continuous convolution operators
  for visual tracking.
\newblock In {\em European Conference on Computer Vision}, pages 472--488.
  Springer, 2016.

\bibitem{faster_rcnn_iccv15}
R.~Girshick.
\newblock Fast r-cnn.
\newblock In {\em International Conference on Computer Vision}, pages
  1440--1448, 2015.

\bibitem{graph_based_vos_cvpr10}
M.~Grundmann, V.~Kwatra, M.~Han, and I.~Essa.
\newblock Efficient hierarchical graph-based video segmentation.
\newblock In {\em 2010 ieee computer society conference on computer vision and
  pattern recognition}, pages 2141--2148. IEEE, 2010.

\bibitem{mask_rcnn_iccv17}
K.~He, G.~Gkioxari, P.~Doll{\'a}r, and R.~Girshick.
\newblock Mask r-cnn.
\newblock In {\em International Conference on Computer Vision}, pages
  2980--2988, 2017.

\bibitem{he2016deep}
K.~He, X.~Zhang, S.~Ren, and J.~Sun.
\newblock Deep residual learning for image recognition.
\newblock In {\em IEEE Conference on Computer Vision and Pattern Recognition},
  pages 770--778, 2016.

\bibitem{goturn_eccv16}
D.~Held, S.~Thrun, and S.~Savarese.
\newblock Learning to track at 100 fps with deep regression networks.
\newblock In {\em European Conference on Computer Vision}, pages 749--765.
  Springer, 2016.

\bibitem{kcf_pami15}
J.~F. Henriques, R.~Caseiro, P.~Martins, and J.~Batista.
\newblock High-speed tracking with kernelized correlation filters.
\newblock {\em IEEE transactions on pattern analysis and machine intelligence},
  37(3):583--596, 2015.

\bibitem{motion_guided_vos_cvpr18}
P.~Hu, G.~Wang, X.~Kong, J.~Kuen, and Y.-P. Tan.
\newblock Motion-guided cascaded refinement network for video object
  segmentation.
\newblock In {\em International Conference on Computer Vision and Pattern
  Recognition}, pages 1400--1409, 2018.

\bibitem{videomatch_eccv18}
Y.-T. Hu, J.-B. Huang, and A.~G. Schwing.
\newblock Videomatch: Matching based video object segmentation.
\newblock In {\em European Conference on Computer Vision}, pages 56--73, 2018.

\bibitem{appearance_model_for_end2end_vos_arxiv1811}
J.~Johnander, M.~Danelljan, E.~Brissman, F.~S. Khan, and M.~Felsberg.
\newblock A generative appearance model for end-to-end video object
  segmentation.
\newblock {\em arXiv preprint arXiv:1811.11611}, 2018.

\bibitem{rt_mdnet_eccv18}
I.~Jung, J.~Son, M.~Baek, and B.~Han.
\newblock Real-time mdnet.
\newblock In {\em European Conference on Computer Vision}, pages 89--104, 2018.

\bibitem{mask_rcnn_oral_pdf_iccv17}
P.~D. Kaiming~He, Georgia~Gkioxari and R.~Girshick.
\newblock Mask r-cnn: A perspective on equivariance.
\newblock
  \url{http://kaiminghe.com/iccv17tutorial/maskrcnn_iccv2017_tutorial_kaiminghe.pdf}.
\newblock Accessed 3, 2017.

\bibitem{video_det_tubelet_proposal_cvpr17}
K.~Kang, H.~Li, T.~Xiao, W.~Ouyang, J.~Yan, X.~Liu, and X.~Wang.
\newblock Object detection in videos with tubelet proposal networks.
\newblock In {\em CVPR}, 2017.

\bibitem{lucid_datadreaming_arxiv18}
A.~Khoreva, R.~Benenson, E.~Ilg, T.~Brox, and B.~Schiele.
\newblock Lucid data dreaming for video object segmentation.
\newblock 2018.

\bibitem{kingma2014adam}
D.~P. Kingma and J.~Ba.
\newblock Adam: A method for stochastic optimization.
\newblock {\em arXiv preprint arXiv:1412.6980}, 2014.

\bibitem{vot_pami16}
M.~Kristan, J.~Matas, A.~Leonardis, T.~Vojir, R.~Pflugfelder, G.~Fernandez,
  G.~Nebehay, F.~Porikli, and L.~\v{C}ehovin.
\newblock A novel performance evaluation methodology for single-target
  trackers.
\newblock {\em IEEE Transactions on Pattern Analysis and Machine Intelligence},
  38(11):2137--2155, Nov 2016.

\bibitem{siamrpn_cvpr18}
B.~Li, J.~Yan, W.~Wu, Z.~Zhu, and X.~Hu.
\newblock High performance visual tracking with siamese region proposal
  network.
\newblock In {\em Proceedings of the IEEE Conference on Computer Vision and
  Pattern Recognition}, pages 8971--8980, 2018.

\bibitem{dyenet_eccv18}
X.~Li and C.~Change~Loy.
\newblock Video object segmentation with joint re-identification and
  attention-aware mask propagation.
\newblock In {\em European Conference on Computer Vision}, pages 93--110, 2018.

\bibitem{lin2014microsoft}
T.-Y. Lin, M.~Maire, S.~Belongie, J.~Hays, P.~Perona, D.~Ramanan,
  P.~Doll{\'a}r, and C.~L. Zitnick.
\newblock Microsoft coco: Common objects in context.
\newblock In {\em European Cconference on Computer Vision}, pages 740--755,
  2014.

\bibitem{prevmvos_accv18}
J.~Luiten, P.~Voigtlaender, and B.~Leibe.
\newblock Premvos: Proposal-generation, refinement and merging for video object
  segmentation.
\newblock In {\em Asian Conference on Computer Vision}, 2018.

\bibitem{vos_wo_tempinfo_pami18}
K.-K. Maninis, S.~Caelles, Y.~Chen, J.~Pont-Tuset, L.~Leal-Taix\'e, D.~Cremers,
  and L.~{Van Gool}.
\newblock Video object segmentation without temporal information.
\newblock {\em IEEE Transactions on Pattern Analysis and Machine Intelligence
  (TPAMI)}, 2018.

\bibitem{bilateral_space_vos_cvpr16}
N.~M{\"a}rki, F.~Perazzi, O.~Wang, and A.~Sorkine-Hornung.
\newblock Bilateral space video segmentation.
\newblock In {\em IEEE Conference on Computer Vision and Pattern Recognition},
  pages 743--751, 2016.

\bibitem{mdnet_cvpr16}
H.~Nam and B.~Han.
\newblock Learning multi-domain convolutional neural networks for visual
  tracking.
\newblock In {\em IEEE Conference on Computer Vision and Pattern Recognition},
  pages 4293--4302, 2016.

\bibitem{peng2017large}
C.~Peng, X.~Zhang, G.~Yu, G.~Luo, and J.~Sun.
\newblock Large kernel matters—improve semantic segmentation by global
  convolutional network.
\newblock In {\em IEEE Conference on Computer Vision and Pattern Recognition},
  pages 1743--1751, 2017.

\bibitem{masktrack_cvpr17}
F.~Perazzi, A.~Khoreva, R.~Benenson, B.~Schiele, and A.~Sorkine-Hornung.
\newblock Learning video object segmentation from static images.
\newblock In {\em IEEE Conference on Computer Vision and Pattern Recognition},
  pages 3491--3500, 2017.

\bibitem{davis_eval_cvpr16}
F.~Perazzi, J.~Pont-Tuset, B.~McWilliams, L.~Van~Gool, M.~Gross, and
  A.~Sorkine-Hornung.
\newblock A benchmark dataset and evaluation methodology for video object
  segmentation.
\newblock In {\em IEEE Conference on Computer Vision and Pattern Recognition},
  pages 724--732, 2016.

\bibitem{fully_connected_vos_iccv15}
F.~Perazzi, O.~Wang, M.~Gross, and A.~Sorkine-Hornung.
\newblock Fully connected object proposals for video segmentation.
\newblock In {\em Proceedings of the IEEE international conference on computer
  vision}, pages 3227--3234, 2015.

\bibitem{refinement_module_eccv16}
P.~O. Pinheiro, T.-Y. Lin, R.~Collobert, and P.~Doll{\'a}r.
\newblock Learning to refine object segments.
\newblock In {\em European Conference on Computer Vision}, pages 75--91.
  Springer, 2016.

\bibitem{davis17_arxiv}
J.~Pont-Tuset, F.~Perazzi, S.~Caelles, P.~Arbel{\'a}ez, A.~Sorkine-Hornung, and
  L.~Van~Gool.
\newblock The 2017 davis challenge on video object segmentation.
\newblock {\em arXiv preprint arXiv:1704.00675}, 2017.

\bibitem{davis17_arxiv17}
J.~Pont-Tuset, F.~Perazzi, S.~Caelles, P.~Arbel{\'a}ez, A.~Sorkine-Hornung, and
  L.~Van~Gool.
\newblock The 2017 davis challenge on video object segmentation.
\newblock {\em arXiv preprint arXiv:1704.00675}, 2017.

\bibitem{onavos_bmvc17}
P.~Voigtlaender and B.~Leibe.
\newblock Online adaptation of convolutional neural networks for video object
  segmentation.
\newblock In {\em British Machine Vision Conference}, 2017.

\bibitem{otb2013}
Y.~Wu, J.~Lim, and M.-H. Yang.
\newblock Online object tracking: A benchmark.
\newblock In {\em Proceedings of the IEEE conference on computer vision and
  pattern recognition}, pages 2411--2418, 2013.

\bibitem{rgmp_cvpr18}
S.~Wug~Oh, J.-Y. Lee, K.~Sunkavalli, and S.~Joo~Kim.
\newblock Fast video object segmentation by reference-guided mask propagation.
\newblock In {\em IEEE Conference on Computer Vision and Pattern Recognition},
  pages 7376--7385, 2018.

\bibitem{monet_cvpr18}
H.~Xiao, J.~Feng, G.~Lin, Y.~Liu, and M.~Zhang.
\newblock Monet: Deep motion exploitation for video object segmentation.
\newblock In {\em International Conference on Computer Vision and Pattern
  Recognition}, pages 1140--1148, 2018.

\bibitem{Xie2016}
S.~Xie, R.~Girshick, P.~Doll{\'a}r, Z.~Tu, and K.~He.
\newblock Aggregated residual transformations for deep neural networks.
\newblock In {\em IEEE Conference on Computer Vision and Pattern Recognition},
  pages 5987--5995, 2017.

\bibitem{yvos_eccv18}
N.~Xu, L.~Yang, Y.~Fan, J.~Yang, D.~Yue, Y.~Liang, B.~Price, S.~Cohen, and
  T.~Huang.
\newblock Youtube-vos: Sequence-to-sequence video object segmentation.
\newblock In {\em European Conference on Computer Vision}, pages 603--619,
  2018.

\bibitem{Yang2018osmn}
L.~Yang, Y.~Wang, X.~Xiong, J.~Yang, and A.~K. Katsaggelos.
\newblock Efficient video object segmentation via network modulation.
\newblock In {\em IEEE Conference on Computer Vision and Pattern Recognition},
  pages 6499--6507, 2018.

\end{thebibliography}
}

\end{document}